\documentclass[journal]{IEEEtran}

\ifCLASSINFOpdf \else
   \usepackage[dvips]{graphicx}
\fi
\usepackage{url}
\usepackage{multicol}
\usepackage{graphicx}
\usepackage{subfig}
\usepackage{algorithm,algpseudocode}
\usepackage{mathdots}
\usepackage{graphics}
\usepackage{multirow,diagbox}
\usepackage{tabularx}
\usepackage{paralist}
\usepackage{verbatim}
\usepackage{booktabs}
\usepackage{amsfonts}
\usepackage{amsmath}
\usepackage{placeins}
\usepackage{bm}
\usepackage{epsfig}
\usepackage{threeparttable}
\usepackage{setspace}
\usepackage{float}

\begin{document}

\title{Learning unbiased zero-shot semantic segmentation networks via transductive transfer}

\author{Haiyang~Liu, Yichen~Wang, Jiayi~Zhao, Guowu Yang, and Fengmao~Lv
  \thanks{This work is supported by the National Natural Science Foundation of China (No. 11829101 and 11931014) and the Fundamental Research Funds for the Central Universities of China  (No. JBK1806002).  H. Liu and Y. Wang contribute equally to this work. (Corresponding author: Fengmao Lv.)}
    \thanks{H. Liu, Y. Wang and G. Yang are with School of Computer Science and Engineering, University of Electronic Science and Technology of China, Chengdu, 611731, China.}
    \thanks{J. Zhao and F. Lv are with Center of Statistical Research \& School of Statistics, Southwestern University of Finance and Economics, China (e-mail: fengmaolv@126.com).}
}
\markboth{} {Khan
\MakeLowercase{\textit{et al.}}: Detection and Localization of
Superimposed Scorebox in Broadcast Sports Videos} \maketitle

\begin{abstract}
Semantic segmentation, which aims to acquire a detailed understanding of images, is an essential issue in computer vision. However, in practical scenarios, new categories that are different from the categories in training usually appear. Since it is impractical to collect labeled data for all categories, how to conduct zero-shot learning in semantic segmentation establishes an important problem. Although the attribute embedding of categories can promote effective knowledge transfer across different categories, the prediction of segmentation network reveals obvious bias to seen categories. In this paper, we propose an easy-to-implement transductive approach to alleviate the prediction bias in zero-shot semantic segmentation. Our method assumes that both the source images with full pixel-level labels and unlabeled target images are available during training.  To be specific, the source images are used to learn the relationship between visual images and semantic embeddings, while the target images are used to alleviate the prediction bias towards seen categories. We conduct comprehensive experiments on diverse split s of the PASCAL dataset. The experimental results clearly demonstrate the effectiveness of our method.
\end{abstract}


\IEEEpeerreviewmaketitle

\section{Introduction}
\label{sec:Introduction}

\IEEEPARstart{W}{ith} the rapid development of deep learning, semantic segmentation has achieved great advances over the recent years. To train a good semantic segmentation model, we need a large amount of images with full pixel-level labels. However, it is impractical to collect labeled data for all categories~\cite{DBLP:conf/cvpr/ArdeshirCS15,DBLP:conf/cvpr/LiuH15,DBLP:conf/cvpr/BanicaS15}. In practical scenarios, new categories that are different from the categories in training usually appear.  In this case, semantic segmentation neural networks struggle to make correct predictions for them.

\begin{figure}[t]\centering
     \subfloat[]{\includegraphics[width=1.0in]{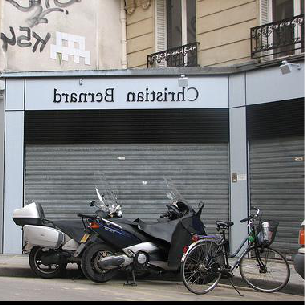}}\hspace{0.3mm}
     \subfloat[]{\includegraphics[width=1.0in]{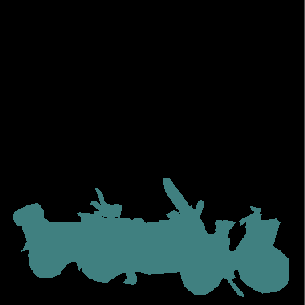}}\hspace{0.3mm}
     \subfloat[]{\includegraphics[width=1.0in]{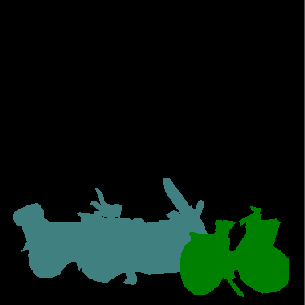}}
     \caption{(a)  The original image that contains bicycle and
motorcycle. The motorcycle class is included in the training data and bicycle is unseen during training. (b) The incorrect semantic segmentation results caused by the problem of prediction bias. (c) The pixel-level ground truth.} \label{intro}
\end{figure}

Zero-Shot Learning (ZSL) is mainly studied in the context of image recognition. It aims to recognize unseen (target) categories by transferring knowledge from seen (source) categories~\cite{DBLP:conf/cvpr/KodirovXG17,DBLP:conf/cvpr/XianLSA18,DBLP:conf/cvpr/PaulKM19}. For zero-shot semantic segmentation, the previous methods propose to incorporate semantic embeddings of categories to semantic segmentation neural networks~\cite{DBLP:conf/cvpr/XianC0SA19}.

However, as shown in Fig. \ref{intro},  the existing zero-shot semantic segmentation approaches reveal strong bias towards seen categories in Generalized ZSL(GZSL)~\cite{DBLP:conf/cvpr/SongSYLS18}. Since the relationship between visual images and semantic embeddings is entirely captured over the source data, the visual instances will show obvious bias to source categories. Hence, the semantic segmentation model tends to map the pixels of unseen images to  the semantic embeddings of source categories and obtains poor generalization performance~\cite{DBLP:journals/pami/ChenPKMY18}. In this paper, we propose an easy-to-implement transductive approach to alleviate the prediction bias in zero-shot semantic segmentation~\cite{DBLP:conf/cvpr/LongSD15}. Our method assumes that both the source images with full pixel-level labels and unlabeled target images are available during training. It should be noted that our method is also useful for the conventional ZSL setting since it can regularize the segmentation model to make more reasonable predictions for target images, which  produces gradients to make the semantic segmentation neural network better fit them~\cite{DBLP:journals/corr/abs-1906-00817}.

Overall, the main contribution of this work is three-fold:
\begin{itemize}
\vspace{0.1cm}
\item We propose to introduce unlabeled target images in zero-shot semantic segmentation to alleviate the prediction bias towards source categories. To the best of our knowledge, this is the first work to conduct transductive learning in zero-shot semantic segmentation.
\vspace{0.1cm}
\item We propose to alleviate the prediction bias through explicitly projecting the pixels of target images to other points in the semantic embedding space. Our proposed method is both effective and easy-to-implement.
\vspace{0.1cm}
\item We conduct comprehensive experiments on diverse splits of the PASCAL dataset. The experimental results clearly demonstrate the effectiveness of our method.
\end{itemize}

\begin{figure*}[h] \centering
   \epsfig{figure=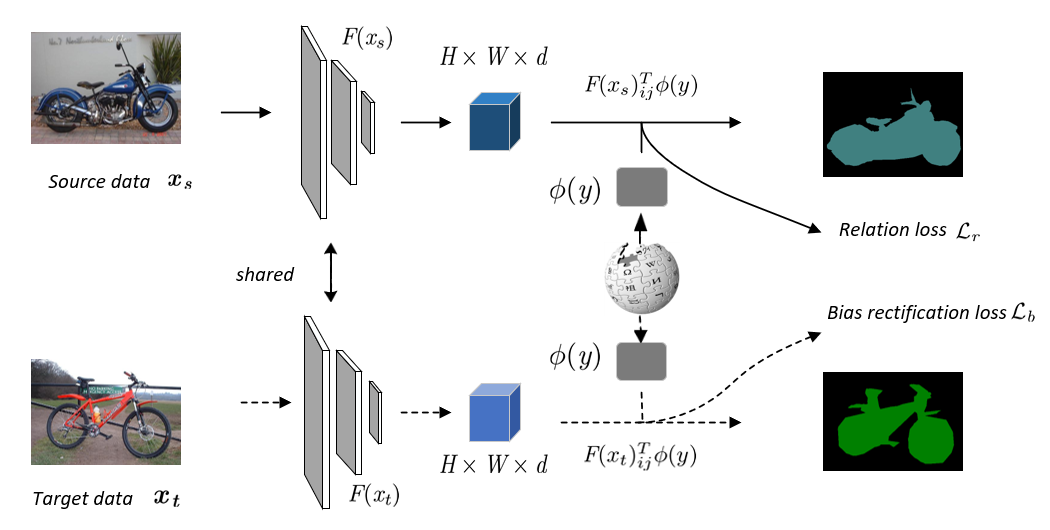, width=0.87\textwidth}
   \caption{The overall architecture of our proposed method.}
   \label{fig:ours}
\end{figure*}

\section{Problem statement \& motivation}
\label{sec:methodology}
Denote by  $\bm{x}_s \in \mathbb{R}^{H \times W \times 3}$ the source images and $\bm{x}_t \in \mathbb{R}^{H \times W \times 3}$ the target images, where $H$ and $W$ are the  height and width, respectively. In zero-shot semantic segmentation, the source and target images have pixel-level labels of different categories, with $\bm{y}_s \in \mathcal{Y}_t^{ H \times W }$, $\bm{y}_t \in \mathcal{Y}_t^{ H \times W}$ and $\mathcal{Y}_s  \bigcap \mathcal{Y}_t = \o$. Like in zero-shot image recognition, the semantic embeddings of categories denoted by $\phi(y) \in \mathbb{R}^{d}$, 
where $y \in \mathcal{Y}_s  \bigcup \mathcal{Y}_t$, provides the effective information to bridge the source and target domains by projecting the visual pixels into the shared semantic space. This work proposes to introduce unlabeled target images in zero-shot semantic segmentation to alleviate the prediction bias towards seen categories. To be specific, we assume that both  labeled source images $\mathcal{D}_s=\{(\bm{x}_s,\bm{y}_s) \}_{s \in \mathcal{S}} $ and unlabeled target images $\mathcal{D}_t=\{(\bm{x}_t)  \}_{t \in \mathcal{T}}$  are available for training, with the goal of obtaining good pixel-level prediction for target images in the conventional setting or for both source and target images in the generalized setting.

A crucial point lies in how to leverage the unlabeled target images in zero-shot semantic segmentation. For this, a natural idea is to perform self-training on the unlabeled target images. However, we argue that directly conducting self-training is not suitable in zero-shot learning. The performance of self-training heavily depends on the initial weights. As indicated above, since the segmentation network has obvious prediction bias on target images, self-training will produce poor pseudo labels for them and cause the problem of negative transfer. To tackle this issue, our method proposes to  project the pixels of target images to the region specified by all target categories in the semantic embedding space, instead of one single point specified by the pseudo label. Our method can effectively alleviate the prediction bias towards seen categories, but not incur the risk of negative transfer.

\section{Approach}

The overall architecture of our model is shown in Figure 2.  The source images are used to learn the relationship between visual images and semantic embeddings, while the target images are used to alleviate the prediction bias towards seen categories.

\subsection{Build visual-semantic relation}

To construct the relationship  between visual images and semantic information, our method uses a FCN backbone to project the pixel of input images into the semantic space shared by both source and target classes: $F(\bm{x}) \in \mathbb{R}^{H \times W \times d} $. The inner product between the visual feature of each pixel $(i,j)$ and the semantic embedding of each category can reveal the relation score: $F(\bm{x})_{ij}^{T}\phi({y})$. To learn the network that can capture the visual-semantic relations, we normalize the relation score by a softmax operation:
\begin{equation}
p(\bm{\hat{y}}_{ij}=y|\bm{x}_{ij}) = \frac{F(\bm{x})_{ij}^{T}\phi({y})}{\sum_{y \in \mathcal{Y}_s \bigcup \mathcal{Y}_t}^{} F(\bm{x})_{ij}^{T}\phi({y})}, \nonumber
\end{equation}
and  train the network with the standard cross-entropy loss:
\begin{equation}
\mathcal{L}_r = \sum_{s \in \mathcal{S}} \mathcal{L}_{seg} (\bm{y}_s, \bm{\hat y}_s), \nonumber
\end{equation}
where $\mathcal{L}_{seg}$ is the pixel-wise cross-entropy loss. Since the target domain does not have  labels, $\mathcal{L}_{seg}$ is trained with only the source images.

\subsection{Alleviate prediction bias}
To alleviate the problem of projection bias towards source classes, our method proposes to project the pixel of target images to the region specified by  target classes in the semantic embedding space, which can be implemented by encouraging the pixel of target images to have large probabilities of belonging to any target class:
\begin{equation}
\mathcal{L}_{{b}}= - \sum_{t \in \mathcal{T}}  \sum_{i=1}^{H} \sum_{j=1}^{W} \ln \sum_{k \in \mathcal{Y}_{t}} p( \bm{\hat{y}}_{ij}=k | \bm{x}_t ), \nonumber
\end{equation}
where $p( \bm{\hat{y}}_{ij}=k | \bm{x}_t )$ represents the  probability that the pixel $(i,j)$ of the target image $\bm{x}_t$ belongs to class $k$.  Compared with self-training, our method does not rely on producing pseudo labels to rectify the prediction bias and hence can avoid the negative transfer caused by incorrect pseudo labels. It should be noted that our method is also useful for the conventional ZSL setting since it can regularize the segmentation model to make more reasonable predictions for target images, which produces gradients to make the semantic segmentation neural network better fit them.

\subsection{Model overview}

To sum up, with the above sub-objectives, our final objective function is formulated as follows:
\begin{equation}
\min \mathcal{L}_{{r}}+\lambda_{{}} \mathcal{L}_{{b}}, \nonumber
\end{equation}
where $\lambda$ is the hyper-parameter that balances the importance of the bias rectification loss $\mathcal{L}_{{b}}$.

$\mathcal{L}_b$ encourages the network to better fit the target images  by explicitly rectifying the prediction bias towards source classes. With better network weights to map the pixel of target images into the semantic space, self-training has the potential to produce better pseudo labels. Hence, after training the segmentation network with $\mathcal{L}_{{b}}$, we can progressively conduct self-training over the target images.

\section{Experiments}
\label{sec:exp}

Following \cite{DBLP:conf/cvpr/XianC0SA19}, we use the concatenation of two different word vectors, i.e. word2vec trained on Google News~\cite{JoulinFastText} and fastText trained on Common Crawl~\cite{JoulinFastText}, to construct the semantic space shared by source and target classes.  We adopt the DeepLabv2   with VGG-16 backboneïœ\cite{DBLP:journals/corr/ChenPK0Y16}, which is pre-trained on ImageNet, to project the pixel of images into the semantic space. Our model is trained by the SGD optimizer with learning rate of $10^{-4}$ and momentum of 0.9.  The mini-batch size is set to 10. In the experiments, we train our model with $\mathcal{L}_r+\lambda \mathcal{L}_b$ for 20 epochs and then conduct self-training over  target images for 3 epochs with the confidence threshold set to 0.6. The learning rate decreases according to the polynomial decay policy with power of 0.9.  The  hyper-parameter $\lambda$ is set to 0.6.

\subsection{Dataset}

In this work, the PASCAL-VOC dataset is used as the benchmark \cite{Everingham10}. PASCAL-VOC consists of 12,031 images with pixel-level labels from 20 categories. Specifically, 5 classes are selected as the  unseen classes, while the remaining 15 classes are used as the seen classes. For each split, the  the unseen classes are shown in Table~\ref{tab:PASCAL}. Following~\cite{DBLP:conf/bmvc/ShabanBLEB17}, we evaluate the performance of our method on the validation set of PASCAL-VOC.

\begin{table}[t]
\renewcommand\arraystretch{1.3}
\fontsize{8.0}{9.0} \selectfont
\small
\caption{The unseen classes in each data split of PASCAL-VOC.} \label{tab:PASCAL}
\begin{tabular}{{
p{0.12\textwidth}|
p{0.30\textwidth}
}}
\toprule[1.5pt]
\hline
\multicolumn{1}{l|}{Dataset} & \multicolumn{1}{l}{Unseen classes}                               \\ \hline
$\mathrm{PASCAL}-5^{\mathrm{0}}$               & aeroplane, bicycle, bird, boat, bottle                            \\
$\mathrm{PASCAL}-5^{\mathrm{1}}$                & bus, car, cat, chair, cow                                         \\
$\mathrm{PASCAL}-5^{\mathrm{2}}$                   & diningtable, dog, horse, motorbike, person                        \\
$\mathrm{PASCAL}-5^{\mathrm{3}}$     &potted plant, sheep, sofa, train, tv/monitor \\ \hline
\end{tabular}
\end{table}

\begin{table*}[t]\centering \scriptsize
\renewcommand\arraystretch{1.05}
\fontsize{7.5}{9.0} \selectfont
\caption{Comparison of different  approaches in the generalized ZSL setting. H is the harmonic mean of the seen classes and unseen classes. The best result is marked in $\textbf{bold}$ font.} \label{tab:gzlss}
\begin
{tabular}
{
m{0.115\textwidth} < {\centering} |
m{0.030\textwidth} < {\centering}
m{0.031\textwidth} < {\centering}
m{0.031\textwidth} < {\centering} |
m{0.030\textwidth} < {\centering}
m{0.030\textwidth} < {\centering}
m{0.030\textwidth} < {\centering} |
m{0.030\textwidth} < {\centering}
m{0.030\textwidth} < {\centering}
m{0.031\textwidth} < {\centering} |
m{0.031\textwidth} < {\centering}
m{0.031\textwidth} < {\centering}
m{0.031\textwidth} < {\centering} |
m{0.031\textwidth} < {\centering}
m{0.030\textwidth} < {\centering}
m{0.030\textwidth} < {\centering}}
\toprule[1.5pt]
\hline
    & \multicolumn{3}{c|}{$\mathrm{PASCAL}-5^{\mathrm{0}}$} & \multicolumn{3}{c|}{$\mathrm{PASCAL}-5^{\mathrm{1}}$} & \multicolumn{3}{c|}{$\mathrm{PASCAL}-5^{\mathrm{2}}$} & \multicolumn{3}{c|}{$\mathrm{PASCAL}-5^{\mathrm{3}}$} & \multicolumn{3}{c}{mean} \\
Method   & seen      & unseen  &H    & seen      & unseen  &H    & seen      & unseen   &H   & seen      & unseen  &H    & seen       & unseen   &H     \\
\hline
SPNet\cite{DBLP:conf/cvpr/XianC0SA19} & 60.1     & 7.0   &12.5     & 55.2     & 25.0   &34.4    & 55.7     & 14.3  &22.8     & 61.6     & 14.2   &23.1    & 58.2     &15.1      & 23.2           \\
ST\cite{DBLP:conf/icml/ZhuL05}      & 63.8     & 1.7  &3.3     & 71.0     & 24.8   &36.7    & $\textbf{66.5}$     & 15.3  &24.9     & 71.3     & 35.3   &47.2    &68.2          & 19.3    &28.0         \\
CBST\cite{DBLP:conf/eccv/ZouYKW18}    & 66.4     & 12.7  &21.4     & 65.3     & 13.4    &22.3   & 65.3     & 10.6  &18.3     & 69.1     & 21.4    &32.7   & 66.5          & 14.5    &23.7        \\
\hline
\textbf{Our method}    & 68.4       & 40.0  &50.5     & 70.4     & 53.5  &60.8     & 63.7     &$\textbf{24.3}$   &$\textbf{35.1}$    & 69.4    & 43.9  &53.8    & 68.0          & 40.4    &50.1        \\
\textbf{Our method+ST}  &$\textbf{69.7}$  &$\textbf{53.5}$  &$\textbf{60.6}$  &$\textbf{71.6}$ 	 &$\textbf{58.7}$ & $\textbf{64.6}$  & 65.9	&19.1 &29.6 &$\textbf{71.4}$	 &$\textbf{53.8}$  &$\textbf{61.3}$  &$\textbf{69.7}$ &$\textbf{46.3}$ &$\textbf{54.0}$ \\
\hline
\end{tabular}
\end{table*}

\begin{table*}[t]\centering \scriptsize
\renewcommand\arraystretch{1.05}
\fontsize{7.5}{9.0} \selectfont
\caption{Comparison of different  approaches in the conventional ZSL setting in terms of mIOU. The best result is marked in $\textbf{bold}$.  Notations â ,  â¡  and  Â§  denote approaches for the one/few-shot setting,  inductive ZSL and  transductive ZSL, respectively.}\label{tab:ZLSS}

\begin{tabular}
{
p{0.03\textwidth}<{\centering}|
p{0.15\textwidth}<{\centering}|
p{0.12\textwidth}<{\centering}
p{0.12\textwidth}<{\centering}
p{0.12\textwidth}<{\centering}
p{0.12\textwidth}<{\centering}|
p{0.12\textwidth}<{\centering}
}
\toprule[1.5pt]
\hline
   &Method   & $\mathrm{PASCAL}-5^{\mathrm{0}}$  & $\mathrm{PASCAL}-5^{\mathrm{1}}$   &$\mathrm{PASCAL}-5^{\mathrm{2}}$   & $\mathrm{PASCAL}-5^{\mathrm{3}}$  & mean \\
\hline
\multirow{4}{*}{â }
&OSLSM \cite{DBLP:conf/bmvc/ShabanBLEB17}  &33.6  &55.3  &40.9   &33.5  &40.8   \\
&co-FCN \cite{DBLP:conf/iclr/RakellySDEL18} &36.7 &50.6  &44.9  &32.4  &41.2   \\
&FG-BG \cite{DBLP:conf/cvpr/CaellesMPLCG17} &27.4 &51.7 &34.0 &26.4 &34.9  \\
&MDL\cite{DBLP:conf/compsac/DongZSZ19} &40.6 &58.5  &$\textbf{47.7}$ &36.6  &45.9\\
\hline
\multirow{2}{*}{â¡}
&SPNet\cite{DBLP:conf/cvpr/XianC0SA19} &27.5  &71.4  &32.9   &47.8  &44.9   \\
&VM\cite{Kato_2019_ICCV} &39.6 &52.6 &41.0 &35.6 &42.2  \\
\hline
\multirow{4}{*}{Â§}
&ST\cite{DBLP:conf/icml/ZhuL05} &2.0  &40.7  &18.5   &49.4  &27.7   \\
&CBST\cite{DBLP:conf/eccv/ZouYKW18} &19.6  &25.8  &15.4   &43.8  &26.2   \\
&\textbf{Our method}  &58.4  &74.1 &{41.2} &65.9 &59.9 \\
&\textbf{Our method+ST}  &$\textbf{65.7}$ & $\textbf{87.8}$  & 26.6   &$\textbf{70.8}$   &$\textbf{62.7}$  \\

\hline
\end{tabular}
\end{table*}

\subsection{Baseline}
We compare the performance of the proposed method with the following baselines:
\begin{itemize}
\item  \textbf{One-shot baselines} include One-shot Learning for Semantic Segmentation (OSLSM) from \cite{DBLP:conf/bmvc/ShabanBLEB17}, and conditional Fully Convolutional Networks (co-FCN) from \cite{DBLP:conf/iclr/RakellySDEL18}.
\item  \textbf{Few-shot baselines} include Foreground-Background (FG-BG) from \cite{DBLP:conf/cvpr/CaellesMPLCG17}, and Multi-scale Discriminative Location-aware network (MDL) \cite{DBLP:conf/compsac/DongZSZ19}.
\item  \textbf{Inductive baselines} include Semantic Projecting Network (SPNet) from \cite{DBLP:conf/cvpr/XianC0SA19} and Variational Mapping (VM) from \cite{Kato_2019_ICCV}.
\item  \textbf{Transductive baselines} include Self-Training (ST) from \cite{DBLP:conf/icml/ZhuL05} and Class-Balanced Self-Training (CBST) from \cite{DBLP:conf/eccv/ZouYKW18}.
\end{itemize}
Note that all the baselines use VGG-16 as the backbone network.

\begin{figure*}[t]\centering    {
    \includegraphics[width=1.07in]{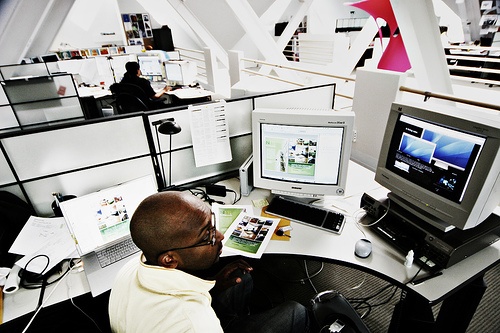}}\hspace{0.3mm}
    {\includegraphics[width=1.07in]{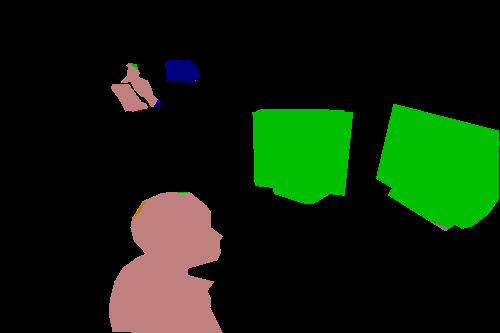}}\hspace{0.3mm}
    {\includegraphics[width=1.07in]{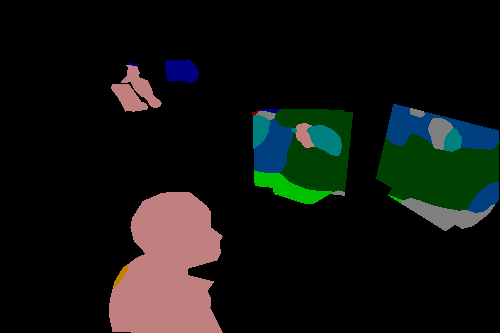}}\hspace{0.3mm}
    {\includegraphics[width=1.07in]{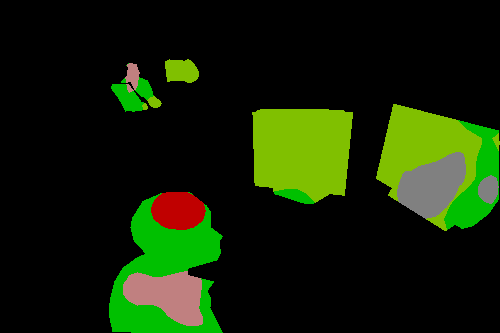}}\hspace{0.3mm}
    {\includegraphics[width=1.07in]{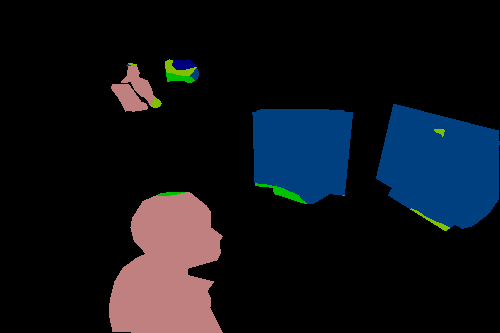}}\hspace{0.3mm}
    {\includegraphics[width=1.07in]{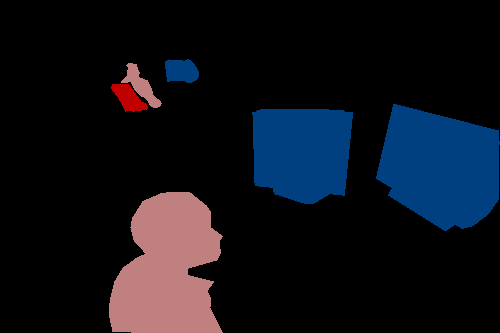}}\\
    \begin{spacing}{1.3}
    \end{spacing}
    {\includegraphics[width=1.07in]{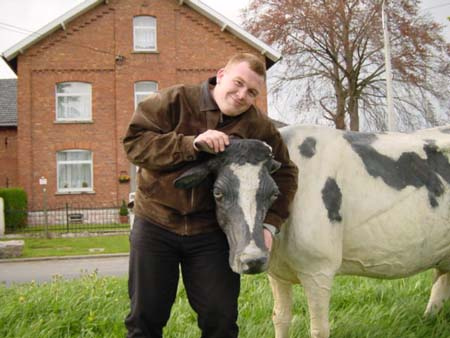}}\hspace{0.3mm}
    {\includegraphics[width=1.07in]{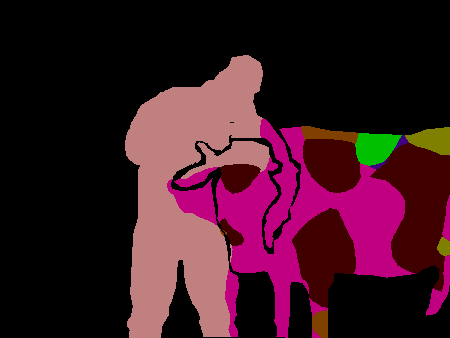}}\hspace{0.3mm}
    {\includegraphics[width=1.07in]{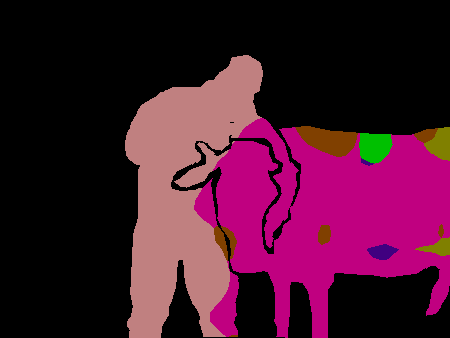}}\hspace{0.3mm}
    {\includegraphics[width=1.07in]{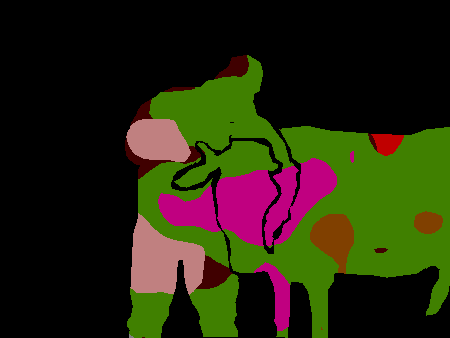}}\hspace{0.3mm}
    {\includegraphics[width=1.07in]{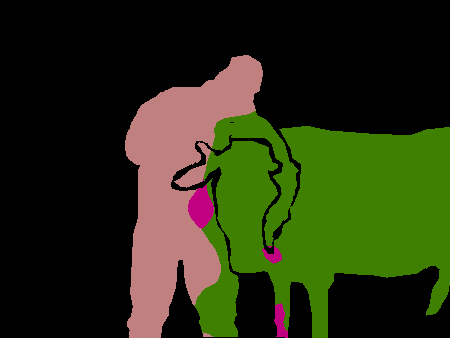}}\hspace{0.3mm}
    {\includegraphics[width=1.07in]{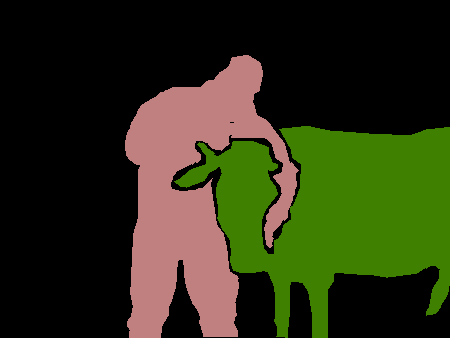}}\\
    \begin{spacing}{1.3}
    \end{spacing}
    \subfloat[image]{\includegraphics[width=1.07in]{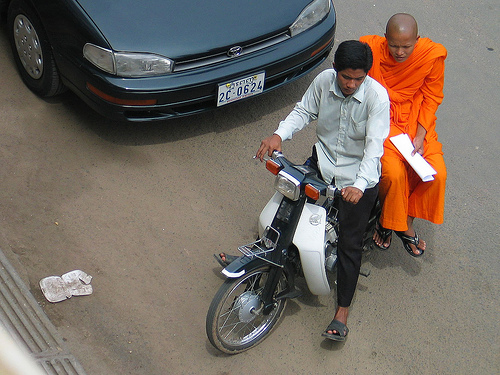}}\hspace{0.3mm}
    \subfloat[CBST]{\includegraphics[width=1.07in]{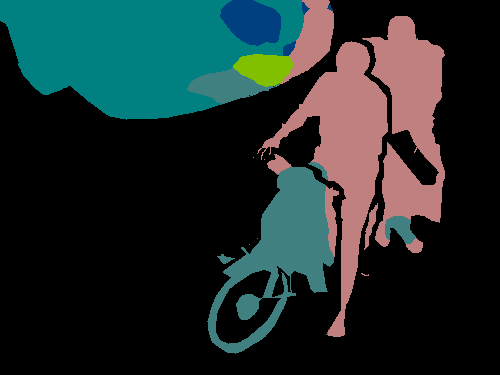}}\hspace{0.3mm}
    \subfloat[ST]{\includegraphics[width=1.07in]{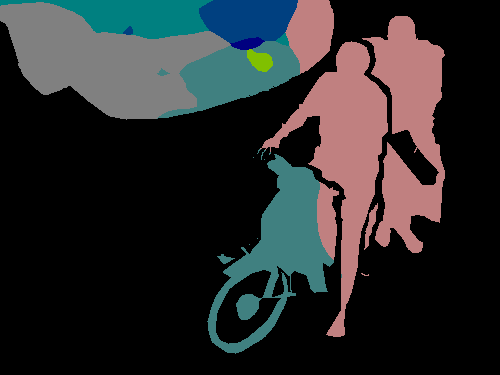}}\hspace{0.3mm}
    \subfloat[SPNet]{\includegraphics[width=1.07in]{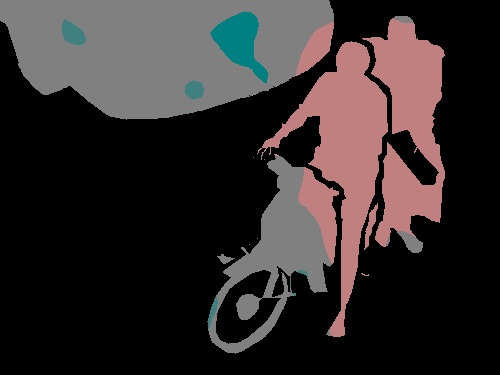}}\hspace{0.3mm}
    \subfloat[our method]{\includegraphics[width=1.07in]{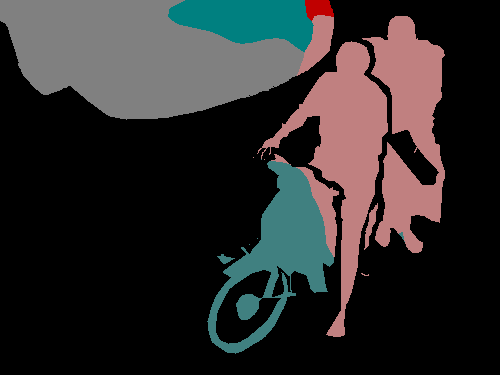}}\hspace{0.3mm}
    \subfloat[ground truth]{\includegraphics[width=1.07in]{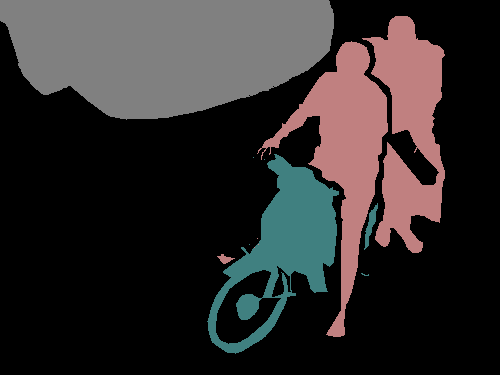}}
     \caption{The qualitative results for  $\mathrm{PASCAL}-5^{\mathrm{3}}$ in the generalized ZSL setting. (a) the input images. (b, c, d) the results of the compared baselines. (e) the results of our method. (f) the semantic segmentation ground truth.} \label{visual}
\end{figure*}

\begin{figure}[!h] \centering
  \includegraphics[width=0.475\textwidth, height=0.28\textwidth, trim=0 0 0 0,clip]{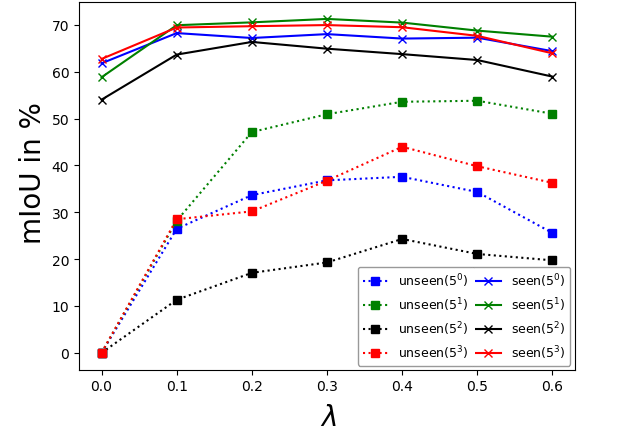}
  \vspace{-0.3cm}
   \caption{The performance of our method by varying $\lambda$.}
   \label{fig:lambda}
\end{figure}

\vspace{0.1cm}

\subsection{Performance comparison}
\noindent$\textbf{Generalized ZSL}$.
Table~\ref{tab:gzlss} displays the comparison of different  mets in the generalized ZSL setting. The mean Intersection-Over-Union(mIOU) of the seen and unseen classes, as well as the harmonic mean of all classes defined in~\cite{DBLP:conf/cvpr/XianC0SA19} are used as the metric of evaluation. Compared with the inductive ZSL baseline SPNet, our method obtains significantly better results by alleviating the prediction bias towards source classes. The performance of both ST and CBST are relatively poor. Since the segmentation network has obvious prediction bias on target images, self-training will produce poor pseudo labels for them and cause the problem of negative transfer. For $\mathrm{PASCAL}-5^{\mathrm{2}}$, progressive self-training decreases the performance. The reason can be that the classes of ``diningtable'' and ``person'' have weak relation with the other classes, and hence the network weights learned by our method are not good enough to produce correct pseudo labels.

\vspace{0.1cm}

\noindent$\textbf{Conventional ZSL}$. Table~\ref{tab:ZLSS} displays the comparison of different  methods in the conventional ZSL setting.  Similar observations can be drawn as in the generalized setting. Our method is also useful for the conventional ZSL setting.
On the split of $\mathrm{PASCAL}-5^{\mathrm{2}}$, the performance of our method is slightly worse than the few-shot baselines co-FCN and MDL. The reason can also be attributed to the weak relation between source and target classes. The few-shot baselines can alleviate this through learning from a few labeled target images.

\subsection{Analysis}

To further examine the effect of the bias rectification loss, we test our model with several different values of the hyper-parameter $\lambda$ on each split. The results are shown in the Figure~\ref{fig:lambda}. We observe that when we increase the value of $\lambda$ from 0 to 0.3, the mIoU of both seen and unseen classes are significantly improved. This observation clearly demonstrates the effectiveness of the proposed bias rectification loss. When the value of $\lambda$ is too large, we can observe a significant performance drop for both seen and unseen classes. This is because that the network focuses too much on alleviating the prediction bias, but ignores to build the relationship between visual images and semantic embeddings.

\section{Conclusion}
\label{sec:con}

This work proposes an easy-to-implement transductive approach to alleviate the prediction bias in zero-shot semantic segmentation. In our method, the source images are used to learn the relationship between visual images and semantic embeddings through projecting the pixels of source images to points specified by the corresponding source categories in the  semantic embedding space, while the target images are used to alleviate the prediction bias towards seen categories through projecting the pixels of target images to other points in the semantic embedding space. The experimental results clearly demonstrate the effectiveness of our method.

\bibliographystyle{IEEEtran}
\bibliography{IEEEtran}

\end{document}